\documentclass[journal]{IEEEtran}
\IEEEoverridecommandlockouts

\usepackage{cite}
\usepackage{amsmath,amssymb,amsfonts}
\usepackage{graphicx}
\usepackage{textcomp}
\usepackage{xcolor}
\usepackage{hyperref}
\hypersetup{hidelinks}
\usepackage{array}
\usepackage{booktabs}
\usepackage{url}
\usepackage{newtxtext,newtxmath}
\usepackage{tikz}
\usetikzlibrary{positioning, arrows.meta, calc}

\begin{document}

\title{AVD: A Multi-Sensor Teleoperated Wheelchair
Platform with Safety-Aware Control}

\author{Namai~Chandra,
        Madhur~Thareja,
        and~Boby~George,~\IEEEmembership{Senior~Member,~IEEE}%
\thanks{N. Chandra and M. Thareja are with the Indian Institute of
Technology Madras, Chennai 600036, India (e-mail:
23f3000200@es.study.iitm.ac.in; 23f3000178@es.study.iitm.ac.in).}%
\thanks{B. George is with the Department of Electrical Engineering,
Indian Institute of Technology Madras, Chennai 600036, India.}}

\maketitle

\begin{abstract}
Electric wheelchairs predominantly rely on manual joystick interfaces,
which present critical barriers to independent mobility for individuals
with severe motor and visual impairments. This paper presents AVD
(Autonomous Vehicle Drive), a \emph{dependable teleoperation
architecture} for assistive mobility that unifies perception,
communication, embedded control, and remote supervision in a single
low-cost platform built entirely from commercially available hardware.
The central engineering contribution is a safety-oriented design
framework: control authority follows a static hierarchy, a manual
emergency stop and the occupant's joystick pre-empt remote commands,
and the on-board sensing layer has \emph{stop-only} authority. The
compute design partitions primary and safety functions and specifies a
heartbeat-driven safe-state transition for the fault cases considered.
The sensing stack this framework depends on is characterized against
measured data: per-channel acquisition latency, a 2\,025-scan LiDAR
degradation campaign spanning five target configurations and a control under water
mist, airborne dust, and direct aperture occlusion, and attitude
measurements across static inclines to $47^{\circ}$. The lowest pooled
environmental-condition mean retained 76.6\% of the clear-baseline
forward-sector yield, and no scan lost the sector completely. A logged
vibration-induced inertial dropout occurred while LiDAR and camera
logging continued, illustrating complementary channel behavior. The
result is a documented prototype and an explicit validation path for
a deployable wheelchair implementation.
\end{abstract}


\IEEEpeerreviewmaketitle

\section{Introduction}

\begin{figure*}[!t]
\centering
\includegraphics[width=0.85\textwidth]{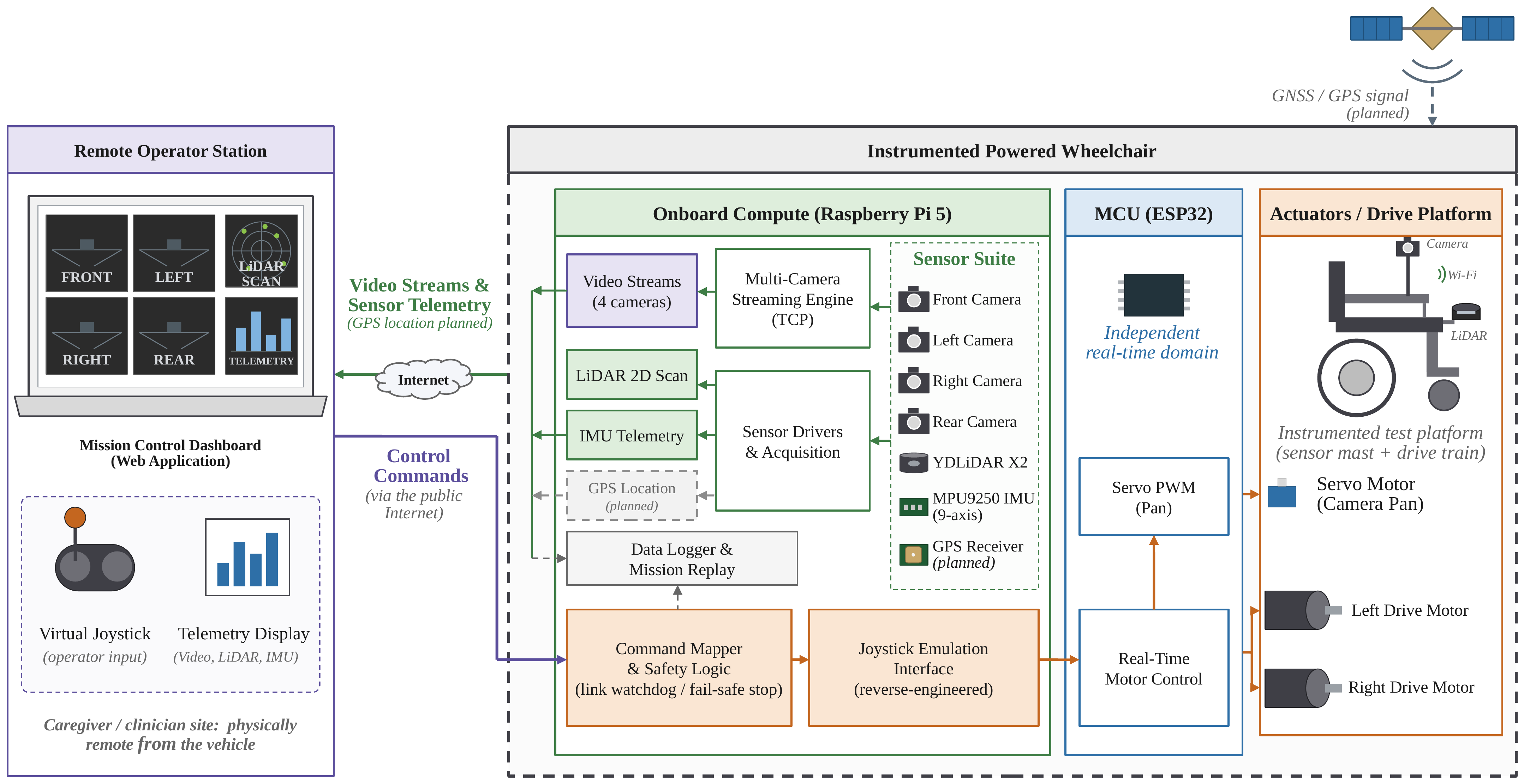}
\caption{Architecture of the AVD platform. The remote operator station (left) is physically remote from the vehicle; everything inside the
right-hand boundary rides on the wheelchair. A single public-internet
link carries video and sensor telemetry outbound to the teleoperator and
control commands inbound to the vehicle. On board, the primary
controller aggregates the sensor suite (cameras, LiDAR, and IMU),
serves the Mission Control
dashboard, records sensor data, and maps teleoperator intent through the
command mapper and safety logic; the microcontroller forms an
independent real-time domain that drives the actuators. Control
authority is strictly ordered: occupant inputs pre-empt the
teleoperator, and the on-board safety layer may only stop the vehicle,
never drive it (Section~\ref{sec:authority}). The GNSS/GPS
receiver and its position path are planned for the field protocol of
Section~\ref{sec:results} and are not yet integrated or validated.}
\label{fig:concept}
\end{figure*}

\IEEEPARstart{M}{obility} is a prerequisite for social participation,
economic inclusion, and quality of life. Approximately 80 million
people worldwide require a wheelchair, while access is estimated at
only 5--35\% depending on the country \cite{who2024}. Among those who do own powered
wheelchairs, the dominant paradigm remains a manual proportional
joystick, a modality inaccessible to individuals with quadriplegia,
severe tremors, or profound visual impairment. Simpson
\cite{simpson2005} identified four unmet needs in powered wheelchair
technology: safe navigation in dynamic environments, environmental
awareness, adaptability to residual motor capacity, and affordability.
Two decades later these gaps remain largely unaddressed in
commercially available devices \cite{leaman2017}. Academic smart
wheelchair research has produced sophisticated navigation systems
\cite{levine1999, bourhis1996}, but the distance between laboratory
prototype and deployable product is rarely bridged \cite{garcia2023,
smartwheelchair2018}, in part because of cost, system complexity, and
limited public implementation detail.

Research responses cluster into two groups. The first replaces the
joystick with an alternative input modality; the second pursues
on-board autonomy through SLAM, path planning, and obstacle avoidance,
which demands expensive sensing, dedicated computing, and controlled
environments (Section~\ref{sec:related}). AVD takes a third position.
It treats the remote human as the navigator and asks a narrower
engineering question: what architecture and responses are needed to
drive the vehicle toward a safe state under the fault cases considered?
Fig.~\ref{fig:concept} shows
the resulting architecture, and Section~\ref{sec:arch} develops the
two commitments that follow from the question: stop-only authority for
the on-board sensing layer, and a compute partition that keeps the
element responsible for safety separate from the element most likely
to fail.

\subsection{Contributions}

\begin{itemize}
\item A \textbf{dependable teleoperation architecture} for assistive
mobility, integrating perception, communication, embedded control, and
remote supervision in one system intended for operation in real
environments.
\item A \textbf{fault-tolerant safety framework} employing
hierarchical control authority, watchdog monitoring, communication
supervision, and fail-safe state transitions, with an explicit
response and validation status for the single-fault cases considered
(Section~\ref{sec:ft}, Table~\ref{tab:faults}).
\item An \textbf{experimental characterization methodology} for the
sensing stack such a platform depends on, applied here through
controlled laboratory validation: per-channel acquisition latency, a
2\,025-scan degradation campaign, and attitude characterization to
$47^{\circ}$ (Section~\ref{sec:sensing}). Kilometre-scale field
deployment is the next milestone and is not claimed here.
\item A \textbf{modular, low-cost prototype} on commercially available
hardware, with public sensor-interface software that supports
reproduction and extension toward shared autonomy and remote assistive
care.
\end{itemize}

\section{Related Work}
\label{sec:related}

\begin{table*}[!t]
\caption{Positioning of AVD Against Representative Assistive and
Teleoperated Systems}
\label{tab:related}
\centering
\scriptsize
\renewcommand{\arraystretch}{0.98}
\setlength{\tabcolsep}{3pt}
\begin{tabular}{@{}>{\raggedright\arraybackslash}p{1.0cm}
                  >{\raggedright\arraybackslash}p{2.0cm}
                  >{\raggedright\arraybackslash}p{2.5cm}
                  >{\raggedright\arraybackslash}p{2.2cm}
                  >{\raggedright\arraybackslash}p{3.0cm}
                  >{\raggedright\arraybackslash}p{3.3cm}
                  >{\raggedright\arraybackslash}p{1.1cm}@{}}
\toprule
\textbf{System} & \textbf{Operating paradigm} & \textbf{Sensing
capability} & \textbf{Remote teleop.\ with live feedback} &
\textbf{Safety and fault tolerance} & \textbf{Validation reported} &
\textbf{Open source} \\
\midrule
\cite{multisensor2025} & Assisted autonomy & LiDAR, IMU, odometry,
ultrasonic & None & Obstacle priority over user intent & 100\%
obstacle accuracy, prepared indoor scenes & No \\
\cite{headcontrol2025} & Head-motion control & Head-worn inertial unit
& None & Not reported & $>$90\% directional accuracy at
$\pm45^{\circ}$; 1.87\,km/h & No \\
\cite{lidarwheelchair2024} & Autonomous docking & LiDAR and vision &
None & Not reported & Docking gap below 5\,mm & No \\
\cite{conav2025} & Shared control & LiDAR and depth camera & None &
Shared-control arbitration & Built-environment user trials & No \\
\cite{iotchair2022} & IoT multi-mode control & Gesture, GPS,
inertial, LiDAR, environmental & App control, no live video & Not
reported & Indoor functional evaluation & Not reported \\
\cite{hsu2012} & Multi-mode intelligent wheelchair & Camera and
indoor navigation sensors & Yes, camera feedback & User and caregiver
priority ordering & Functional comparison of five operating modes &
Not reported \\
\cite{aiotframework2025} & Generic robot teleoperation & Single camera
& Yes, single view & Not reported & 0.2\,s control, below 1.2\,s
video, local and international VPN links & Not reported \\
\midrule
\textbf{AVD} & \textbf{Teleoperation-first, safety-aware} &
\textbf{360$^{\circ}$ LiDAR, four cameras, inertial unit} &
\textbf{Yes, multi-camera with obstacle map} & \textbf{Static
authority hierarchy, stop-only layer, heartbeat watchdog,
single-point failure analysis} & \textbf{2\,025-scan degradation
campaign; field trial pending} & \textbf{Partial} \\
\bottomrule
\end{tabular}
\end{table*}

The powered wheelchair joystick, a proportional two-axis analog
controller driving differential speed signals to a motor controller,
has remained essentially unchanged in principle since the earliest
powered-wheelchair control patents \cite{jones1977}, and its reliance
on precise analog manipulation
excludes a large part of the intended population. Alternative-input
research is correspondingly mature. A 2025 study
\cite{headcontrol2025} achieved directional accuracy above 90\% at
$\pm45^{\circ}$ head tilt, demonstrating that commercial wheelchair
motor interfaces can be reached through standardized signalling
without mechanical modification, a principle AVD also relies on.
Eye-gaze control was shown as early as 2015 \cite{eyecontrol2015} and
later combined with voice \cite{eyevoice2020}; neuro-fuzzy voice
control \cite{voice2020} and camera-based gesture recognition that
avoids the cost of depth sensing \cite{handgesture2025} have both been
demonstrated, and brain-computer \cite{bci2025} and electromyographic
\cite{emg2024} interfaces continue to advance, though both require
calibration, specialist hardware, and tolerance of user fatigue. All
of this work improves how intent enters the chair. None of it
addresses what the chair should do when a sensing or communication
element fails.

On the autonomy side, multi-sensor fusion for powered wheelchairs
\cite{multisensor2025} has combined inertial, LiDAR, odometry, and
ultrasonic sensing under Dempster-Shafer evidence theory, reaching
100\% obstacle accuracy in prepared conditions while noting LiDAR
susceptibility to glass reflectance. A 2023 review of multi-sensor
fusion for impaired navigation \cite{sensorfusion2023} categorized
early, middle, and late fusion strategies, with late fusion
generalizing best across environments \cite{deepfusion2024}, and
planar LiDAR remains the workhorse for reactive obstacle detection on
mobile robots \cite{lidarslam2023}. LiDAR has also been used for
millimetre-accurate wheelchair docking \cite{lidarwheelchair2024} and
for shared-control navigation of the built environment
\cite{conav2025}.

Teleoperation latency carries a measurable cognitive cost: a
functional near-infrared spectroscopy study of 41 operators found that
introducing latency raised functional connectivity within and between
the prefrontal and motor cortices and degraded task performance,
evidence of increased operator load rather than merely slower
actuation \cite{latency2024}. Custom TCP transport guarantees in-order,
error-free delivery at the cost of higher latency than UDP, a
trade-off acceptable for low-speed, caregiver-mediated operation
\cite{tcpstream2015}, and adaptive bitrate allocation tuned to
predicted link quality has been demonstrated for teleoperated vehicles
\cite{teleopvideo2021}. A 2025 modular AIoT framework
\cite{aiotframework2025} reported sub-second actuator latency and
video latency below 1.2\,s for web-based robot teleoperation. On the
embedded side, the Raspberry Pi has become a default compute backbone
for assistive wheelchair prototypes \cite{eyecontrol2015, eyevoice2020},
increasingly paired with a networked microcontroller acting as a
dedicated real-time and connectivity domain \cite{iotchair2022,
multimodalnav2026}.

Table~\ref{tab:related} positions AVD on the axes that matter for a
dependable assistive platform. Remote caregiver operation with camera
feedback has previously been demonstrated on an intelligent robotic
wheelchair \cite{hsu2012}, while generic web-based teleoperation has
also been evaluated across distributed networks
\cite{aiotframework2025}. AVD builds on that precedent by combining
four camera channels, LiDAR telemetry, an explicit control-authority
hierarchy, and a safety-controller design. The comparison is a
positioning aid rather than a head-to-head benchmark, since these
systems measure different tasks under different conditions and public
code was counted only when identified in the cited source.

\section{Architecture and Fault-Tolerant Design}
\label{sec:arch}

\subsection{Platform Organization}
\label{sec:platform}

\begin{figure*}[!t]
\centering
\includegraphics[width=0.78\textwidth]{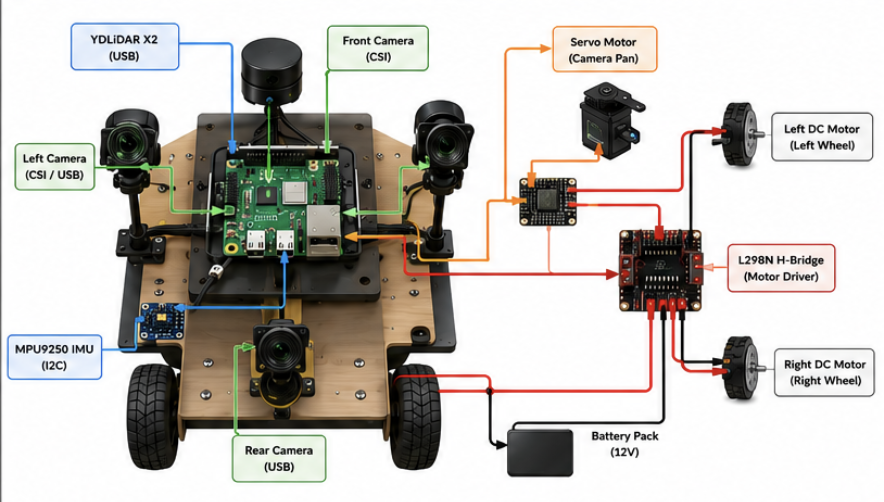}
\caption{Physical realization of the on-board region of
Fig.~\ref{fig:concept} on the proof-of-concept vehicle. The planar
LiDAR, inertial unit, and four camera channels feed the primary
controller; the safety controller drives the servo pan mechanism and
an H-bridge for the left and right wheel motors. All measurements
reported in this paper were collected on this chassis.}
\label{fig:hw}
\end{figure*}

The platform is organized around two computational domains connected
by a serial link, as shown conceptually in Fig.~\ref{fig:concept} and
physically in Fig.~\ref{fig:hw}. The \emph{primary controller}, a
Raspberry~Pi~5, aggregates the sensing channels, runs the multi-camera
streaming engine, serves the teleoperator dashboard, and maps
teleoperator intent into drive commands. The \emph{safety controller}, an ESP32
microcontroller, forms an independent real-time domain that converts
mapped commands into actuation and enforces the safety policy of
Sections~\ref{sec:authority} and~\ref{sec:ft}. Sensing comprises a
YDLiDAR~X2 360$^{\circ}$ planar scanner (8\,m range, 3000
samples/s), four camera channels served at 640$\times$480, and an
inertial unit on a dedicated I$^2$C bus read directly by the primary
controller. Camera streams are prioritized rather than treated
equally: under CPU or network load the peripheral feeds degrade to
10\,fps while the navigation-critical front feed holds 15\,fps.

The present collection logger timestamps the inertial, LiDAR, and
camera channels and writes their samples for offline analysis; the
campaign of Section~\ref{sec:sensing} was collected through this path.
The mission-replay block in Fig.~\ref{fig:concept} is a proposed use of
these logs; an integrated replay tool is not claimed.
Command, drive-state, battery, link-health, and GNSS logging are
requirements for the field protocol of Section~\ref{sec:results}, not
validated features of the current dataset.

Teleoperator intent reaches the drive hardware through a
reverse-engineered joystick interface. Commercial chairs with analog
joystick modules present center-biased differential voltage signals to
the drive controller, and the safety controller synthesizes equivalent
signals at the joystick connector without modifying that controller,
with the intent of preserving its native interlocks. The approach
applies to analog-interface systems such as legacy VR2- and DX-class
controllers. It does not generalize as-is to bus-based systems: the
widely deployed R-Net family carries inter-module traffic over a
CAN-bus protocol rather than analog voltages, so it would require
message emulation rather than signal synthesis, and no universal
cross-vendor control protocol exists \cite{rnet_canbus}.

The proof-of-concept vehicle is a scaled implementation of the sensing
and teleoperation architecture. It drives 12\,V, 2\,A brushed gear
motors through a discrete H-bridge; a target chair may use 24\,V,
15\,A-class motors and a proprietary controller. The prototype
therefore exercises the data and command paths, but it does not
establish equivalent vehicle dynamics, braking, interlock behavior, or
commercial joystick compatibility. Those require bench validation on
the intended wheelchair controller.

\subsection{Control Authority Hierarchy}
\label{sec:authority}

Two sources can request motion, the occupant and the teleoperator; an
on-board safety monitor can only inhibit it. Their relationship is
resolved by a static ordering. In descending precedence:

\begin{enumerate}
\renewcommand{\labelenumi}{P\arabic{enumi}.}
\item \textbf{Manual emergency-stop input.} A button within reach of
the occupant or an attendant drives an interrupt line on the safety
controller and is specified to latch until physically released. It
bypasses the primary controller, link, and dashboard; the present
prototype does not provide a hardwired drive-power cut independent of
the safety controller.
\item \textbf{On-board sensing layer, stop-only.} The sensing channels
feed a safety monitor that runs on-board and is therefore independent
of link health. Its authority is deliberately asymmetric: it may
command a halt, engage the brake, and raise a teleoperator alert, but it
can \emph{never} initiate, resume, or steer motion. It has one verb.
The specified trigger is an obstacle inside the LiDAR close zone
($<$50\,cm) along the direction of travel. Sensor-loss behavior is
handled separately as a degraded-mode design in Section~\ref{sec:ft}.
\item \textbf{Occupant manual override.} Movement of the chair's own
joystick suppresses teleoperated commands immediately and throughout a
configured hold-off. A safety stop remains higher priority, while the
person in the chair is never overruled by the remote operator.
\item \textbf{Teleoperator commands.} Remote drive commands are the
lowest-priority input and are honoured only while fresh.
\end{enumerate}

\subsection{Fault-Tolerant Architecture}
\label{sec:ft}

\begin{figure*}[!t]
\centering
\resizebox{\textwidth}{!}{%
\begin{tikzpicture}[
  font=\footnotesize, >=Stealth,
  blk/.style={draw, rounded corners=2pt, align=center, inner sep=5pt,
              fill=white},
  dom/.style={draw, dashed, thick, rounded corners=4pt, fill=black!3},
  domlbl/.style={font=\footnotesize\bfseries},
  ann/.style={font=\scriptsize, align=center},
  anni/.style={font=\scriptsize\itshape, align=center}
]
\draw[dom] (2.90,4.45) rectangle (7.60,6.85);
\node[domlbl] at (5.25,7.12) {Fault domain A};
\draw[dom] (10.40,4.45) rectangle (15.10,6.85);
\node[domlbl] at (12.75,7.12) {Fault domain B};

\node[blk, text width=1.9cm] (sens) at (1.25,5.65)
     {On-board\\sensing};

\node[blk, text width=4.0cm] (prim) at (5.25,5.65)
     {\textbf{Primary controller}\\[2pt]
      general-purpose OS,\\network stack, media pipeline\\[2pt]
      \emph{non-deterministic}};

\node[blk, text width=4.0cm] (safe) at (12.75,5.65)
     {\textbf{Safety controller}\\[2pt]
      bare-metal loop, no OS,\\no network stack\\[2pt]
      \emph{operational through any\\fault in domain A}};

\node[blk, text width=2.1cm] (drv) at (16.45,5.65)
     {Drive hardware\\\emph{isolated supply}};

\draw[->, thick] (2.55,5.65) -- (2.90,5.65);
\draw[->, very thick] (7.60,5.65) -- (10.40,5.65);
\node[ann, text width=2.5cm, anchor=south] at (9.00,5.82)
     {UART, 115\,200\,baud};
\node[ann, text width=2.5cm, anchor=north] at (9.00,5.50)
     {sequenced frames,\\heartbeat 100\,ms};
\draw[->, thick] (15.10,5.65) -- (15.40,5.65);

\node[blk, text width=1.9cm] (estop) at (11.55,3.45)
     {\textbf{E-stop}\\GPIO interrupt};
\node[blk, text width=2.1cm] (joy) at (14.15,3.45)
     {\textbf{Occupant joystick}\\priority P3};
\draw[->, thick] (11.55,3.95) -- (11.55,4.45);
\draw[->, thick] (14.15,3.95) -- (14.15,4.45);
\node[anni, text width=3.4cm, anchor=east] at (10.30,3.45)
     {hardware paths, independent\\of domain A and of the link};

\node[domlbl, anchor=west] at (3.20,2.30)
     {Escalation on loss of the heartbeat};

\fill[green!25]  (3.40,1.35) rectangle (7.75,1.90);
\fill[orange!30] (7.75,1.35) rectangle (12.58,1.90);
\fill[red!30]    (12.58,1.35) rectangle (15.00,1.90);
\draw (3.40,1.35) rectangle (15.00,1.90);
\draw (7.75,1.35) -- (7.75,1.90);
\draw (12.58,1.35) -- (12.58,1.90);
\node[font=\scriptsize] at (5.57,1.62) {DRIVING};
\node[font=\scriptsize] at (10.16,1.62) {NEUTRAL};
\node[font=\scriptsize] at (13.79,1.62) {BRAKE};

\draw[->] (3.20,0.60) -- (15.30,0.60);
\node[ann, anchor=west] at (15.35,0.60) {$t$};
\foreach \x in {3.40,3.88,4.37,4.85,5.33}{
  \draw[thick] (\x,0.60) -- (\x,1.00);}
\foreach \x in {5.81,6.30,6.78,7.27,7.75}{
  \draw[dotted] (\x,0.60) -- (\x,1.00);
  \node[font=\tiny] at (\x,1.12) {$\times$};}
\node[ann, anchor=east] at (3.15,0.85) {heartbeat};

\draw[dashed] (5.33,0.40) -- (5.33,1.35);
\node[anni, anchor=north] at (5.33,0.35) {link lost};
\draw[dashed] (7.75,0.40) -- (7.75,1.35);
\node[anni, anchor=north] at (7.75,0.35)
     {$T_{\mathrm{wd}}=500$\,ms};
\draw[dashed] (12.58,0.40) -- (12.58,1.35);
\node[anni, anchor=north] at (12.58,0.35)
     {$+\,T_{\mathrm{stop}}=1000$\,ms};
\end{tikzpicture}%
}
\caption{Fault-tolerant system architecture. The safety controller is a
separate compute domain coupled to the primary controller through a
timed command-and-heartbeat link. The specified timeout places the
drive in neutral 500\,ms after the last valid frame and applies the
brake after a further 1000\,ms. The emergency stop and occupant
joystick bypass the primary controller but still depend on the safety
controller. Proposed supply isolation is not implemented on the
present prototype.}
\label{fig:ft}
\end{figure*}
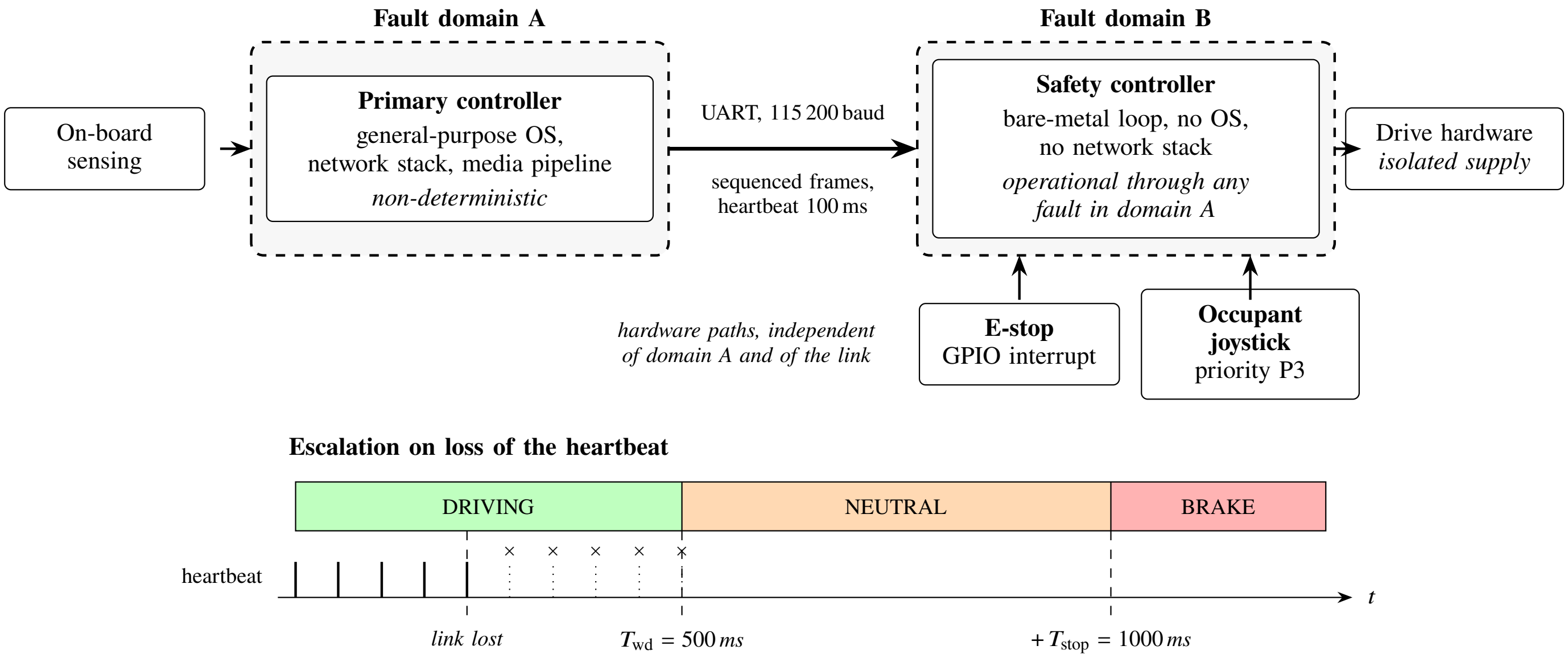

\begin{table*}[!t]
\caption{Single-Fault Design Analysis and Validation Status}
\label{tab:faults}
\centering
\scriptsize
\renewcommand{\arraystretch}{0.98}
\setlength{\tabcolsep}{5pt}
\begin{tabular}{@{}>{\raggedright\arraybackslash}p{4.0cm}
                  >{\raggedright\arraybackslash}p{4.6cm}
                  >{\raggedright\arraybackslash}p{6.4cm}
                  >{\raggedright\arraybackslash}p{2.0cm}@{}}
\toprule
\textbf{Fault} & \textbf{Detection} & \textbf{Safe-state response} &
\textbf{Status} \\
\midrule
Camera channel loss & Frame staleness at primary controller &
Front-feed loss inhibits teleoperation; auxiliary-feed loss raises an
alert; occupant control remains available & Specified \\[1pt]
LiDAR loss & Scan staleness & Remote motion stops; occupant control
remains available; alert raised & Specified \\
Inertial loss & Sample-gap watchdog, $>T_{\mathrm{wd}}$ &
Tip-hazard alerting suspended; teleoperator alerted & Fault observed;
response specified$^{\dagger}$ \\
Link loss or stale command & Sequence number and heartbeat timeout at
safety controller & Neutral at $T_{\mathrm{wd}}=500$\,ms, brake at
$T_{\mathrm{wd}}+T_{\mathrm{stop}}=1500$\,ms & Specified \\
Primary controller crash, hang, or frozen process &
$N_{\mathrm{miss}}=5$ missed 100\,ms heartbeats
($=T_{\mathrm{wd}}$) & Motors disabled, servo centred, brake held,
fault indicated & Specified \\
Safety controller crash or power loss & External watchdog and
normally deasserted motor enable (required) & Drive de-energizes if
enable drops; a frozen-high output remains a design gap & Design gap \\
Occupant joystick disconnect & Continuity and idle-state check &
Teleoperation inhibited; alert raised & Specified \\
Non-critical sensing channel silently below expected rate &
Channel-health monitor against configured minimum rate & Degraded
mode: reduced speed ceiling and alert; full stop on loss of a second
channel &
Specified$^{\ddagger}$ \\
Motor transient or brownout & Isolated logic and drive supplies &
Compute domain unaffected, no reset & Specified$^{\ddagger}$ \\
Any state, manual intervention & GPIO interrupt line & Zero-drive
command latched by safety-controller firmware & Prototype input;
hardwired cut pending \\
\bottomrule
\end{tabular}
\\[3pt]
\raggedright\footnotesize $^{\dagger}$The physical dropout was logged;
the specified alert response was not tested, Section~\ref{sec:attitude}.
$^{\ddagger}$Specified in Section~\ref{sec:ft} but not implemented
on the present vehicle.
\end{table*}

The design objective follows the dependability taxonomy of
\cite{avizienis2004}: faults considered in the design should drive the
vehicle toward a defined safe state. This mirrors the safe-state
principle in functional-safety practice \cite{iec61508} and is relevant
to powered-wheelchair controls covered by \cite{iso7176}; compliance
with either standard is not claimed. The proposed partition separates
the two compute domains (Fig.~\ref{fig:ft}) and assigns a detector,
response, and validation status to each fault case considered
(Table~\ref{tab:faults}). The partition matters
because the two controllers fail in different ways and at different
rates. The primary controller runs a general-purpose operating system
with a network stack, four media processes, and a web service; it can
crash, hang, thrash, or freeze a capture pipeline while remaining
nominally powered. The safety controller runs a single bare-metal loop
with no filesystem and no network stack. Placing the stop authority in
the second domain reduces dependence on the primary controller,
although common power, wiring, and the safety controller itself remain
shared failure points on the prototype.

\emph{(a) Independently timed dead-man watchdog (specified).} The
two domains are coupled by a single 115\,200\,baud UART link carrying
fixed-length command frames, each bearing a monotonic sequence number,
a transmit timestamp, the commanded linear and angular rates, and a
CRC. The primary controller emits a heartbeat on that link at
$f_{\mathrm{hb}}=10$\,Hz ($T_{\mathrm{hb}}=100$\,ms). The design
envelope permits a miss threshold of 3 to 5 beats; the selected design
uses $N_{\mathrm{miss}}=5$ so that the heartbeat threshold coincides
with the command-staleness timeout,
$N_{\mathrm{miss}}\,T_{\mathrm{hb}} = T_{\mathrm{wd}} = 500$\,ms, and
one deadline governs both liveness checks. On expiry the safety
controller commands zero velocity, centres the pan servo, and raises a
fault indication; a further $T_{\mathrm{stop}}=1000$\,ms without a
valid frame engages the brake, 1500\,ms after the last valid frame.
Frames with $\mathrm{seq}\le\mathrm{seq}_{\mathrm{last}}$ are
discarded to reject duplicate or out-of-order commands; deployment
also requires an independently checked maximum command age. After a
timeout, motion remains inhibited until a fresh session is established
and the operator releases and re-engages a dead-man input. Silence
therefore decays to a stop (Fig.~\ref{fig:ft}, lower strip). The P1
interrupt bypasses the primary controller and network, while both the
watchdog and emergency-stop behaviors still rely on correct safety-
controller operation.

\emph{(b) Complementary sensing behavior (measured,
Section~\ref{sec:sensing}).} Rather than fusing the channels into a
single estimate, AVD logs them separately. Section~\ref{sec:sensing}
shows the LiDAR
channel is optical-path limited, degraded by aperture occlusion but
largely invariant to vibration, while the inertial channel is
motion-and-connection limited, degraded by vibration but unaffected by
optical occlusion in the tests. The logged dropout demonstrates one
non-simultaneous failure; it does not establish statistical
independence for channels that share power and primary computation.

\emph{(c) Channel-health monitoring and degraded mode (specified, not
implemented).} Mechanisms (a) and (b) leave one gap: a sensing channel
that goes silent while the command link stays healthy. We therefore
specify a supervisory monitor that tracks each channel's observed
update rate against a configured minimum and
trips when a channel the stop-only layer depends on falls below its
expected rate. This closes exactly the failure logged in
Section~\ref{sec:attitude}, where a dead inertial channel would
otherwise have gone undetected until the next stale-command timeout.
Loss of the navigation-critical front camera or LiDAR inhibits remote
motion. Loss of an auxiliary camera or the inertial channel enters a
\emph{degraded mode} with an alert and reduced remote-command speed;
loss of a second channel commands a stop. Occupant control remains
available below P1/P2 stop authority. Numerical rate, speed, and
recovery thresholds require calibration; the mechanism has not been
built or bench-tested.

\emph{Power isolation and security.} A deployable build powers the
compute domain and the motor drive from separate supplies so that a
motor inrush or stall transient cannot brown out the primary
controller; this is specified but not implemented on the present
vehicle. The proof-of-concept control channel is likewise
unauthenticated, acceptable only on a trusted isolated link, and a
deployable system must carry it over TLS with a per-session credential
validated on the vehicle rather than in the hosted frontend.

\section{Sensor Characterization}
\label{sec:sensing}

A teleoperated mobility device depends on the sensing that feeds its
teleoperator and stop-only layer. This section characterizes acquisition
time, LiDAR behavior under optical degradation, and one observed
inertial-channel dropout.

\subsection{Acquisition Latency and Bandwidth}
\label{sec:latency}

\begin{table}[!t]
\caption{Measured Sensor Acquisition Latency and Bandwidth}
\label{tab:acq}
\centering
\renewcommand{\arraystretch}{0.98}
\setlength{\tabcolsep}{3pt}
\resizebox{\columnwidth}{!}{%
\begin{tabular}{llrrrrrr}
\toprule
\textbf{Channel} & \textbf{Interface} & $N$ & \textbf{Mean} &
\textbf{Med.} & \textbf{SD} & \textbf{Rate} & \textbf{B/w} \\
 & & & (ms) & (ms) & (ms) & (Hz) & (kB/s) \\
\midrule
Inertial & I$^2$C, 400\,kHz & 100 & 6.801 & 6.705 & \textbf{0.256}
& 147.0 & 2.06 \\
LiDAR & USB serial & 30 & 89.744 & 89.674 & \textbf{0.270} & 10.78
& 23.2 \\
Camera & CSI (MIPI) & 20 & 48.028 & 49.871 & 7.075 & 20.8 & 890 \\
\bottomrule
\end{tabular}%
}
\end{table}

Per-channel acquisition latency was measured on-board by timestamping
each read call, with $N$ repetitions per sensor
(Table~\ref{tab:acq}). All values include interpreter and operating
system overhead, which is the latency the application actually
experiences.

The three channels sit in distinct latency regimes. The inertial
channel's 6.80\,ms includes three sensor transactions and user-space
overhead. The LiDAR read call averaged 89.74\,ms, while its reported
mean scan frequency of 10.78\,Hz corresponds to 92.76\,ms per
rotation; these separately recorded quantities should not be treated
as identical. The camera's
48.03\,ms is dominated by image signal processing and JPEG encoding,
not by the CSI bus. Latency \emph{stability} separates the channels
more sharply than latency itself. The inertial and LiDAR channels hold
$\sigma=0.26$ and $0.27$\,ms, whereas the camera's $\sigma=7.08$\,ms
is roughly $27\times$ higher. This is the quantitative argument for
anchoring automatic stop decisions on the LiDAR and inertial channels
rather than on the visually richer but temporally less predictable
camera feed. The bandwidth column is an estimated decoded payload rate
for one instance of each channel, not measured serial or network
traffic; camera streaming remains the dominant load.

\subsection{LiDAR Robustness Under Degradation}
\label{sec:degradation}

\begin{table}[!t]
\caption{LiDAR Forward-Sector ($\pm30^{\circ}$) Robustness Index
(2\,025 Scans, 50 Recordings)}
\label{tab:robust}
\centering
\scriptsize
\renewcommand{\arraystretch}{0.98}
\setlength{\tabcolsep}{3pt}
\begin{tabular}{@{}>{\raggedright\arraybackslash}p{3.5cm}rrrr@{}}
\toprule
 & \textbf{Scans} & \textbf{Index} & \textbf{Raw} &
\textbf{Median} \\
 & & \textbf{(\%)} & \textbf{(\%)} & \textbf{(\%)} \\
\midrule
\multicolumn{5}{@{}l}{\emph{By environmental condition, normalized to
the clear baseline}} \\
Clear baseline (reference)      & 1136 & \textbf{100.0} & 76.58 &
77.78 \\
Water mist                      &  222 & \textbf{95.9}  & 73.45 &
72.73 \\
Airborne dust                   &  116 & \textbf{92.6}  & 70.93 &
70.45 \\
Occluded aperture (film)        &  551 & \textbf{76.6}  & 58.64 &
56.52 \\
\midrule
\multicolumn{5}{@{}l}{\emph{By target configuration, normalized to the
no-obstacle control}} \\
No-obstacle control (reference) &  147 & \textbf{100.0} & 86.02 &
83.67 \\
Retroreflective panel           &  301 & \textbf{90.1}  & 77.50 &
78.26 \\
Matte plastic enclosure         &  581 & \textbf{88.1}  & 75.83 &
77.78 \\
Occluded pair                   &  472 & \textbf{78.9}  & 67.90 &
69.77 \\
Translucent bottle              &  301 & \textbf{75.3}  & 64.78 &
69.77 \\
Crinkled foil                   &  223 & \textbf{63.9}  & 54.97 &
54.55 \\
\bottomrule
\end{tabular}
\\[3pt]
\raggedright\scriptsize Raw is the mean fraction of $\pm30^{\circ}$
sector samples returning a valid range (approximately 43 samples per
scan). Index is that mean expressed as a percentage of the reference
row. Pooled across all 2\,025 scans, complete sector blackout occurred
in 0.00\%.
\end{table}

Because the stop-only layer relies principally on the LiDAR, its
behaviour under adverse optical conditions was characterized
systematically. A labelled campaign of 50 recordings comprising
2\,025 scans sampled five target configurations chosen for their optical
behavior and a no-obstacle control: a matte plastic enclosure
(diffuse, cooperative), a translucent bottle (partial transmission), a
retroreflective panel (strong specular return), a crinkled foil sheet
(chaotic specular scattering), an occluded pair with a second object
hidden from the LiDAR plane but visible to the camera, and a
no-obstacle control. The recording set samples, but does not form a
complete factorial grid across clear baseline, water mist, fine airborne
particulate, and thin-film aperture occlusion at nominal standoffs near
0.10, 0.50, and 1.00\,m. One metadata record is labelled 0.20\,m.
Background structure was deliberately left uncleared, so the data
reflect a cluttered environment rather than a prepared range.

The operationally meaningful metric is not whole-scan return density,
since most of a 360$^{\circ}$ sweep points at open space where a null
return is the \emph{correct} reading. It is the fraction of samples
within the $\pm30^{\circ}$ drive-axis sector, the sector the stop-only
layer actually gates on, that register a valid range. We report this
as a \textbf{robustness index}: each condition's mean valid yield
expressed as a percentage of the reference yield, so the reference is
100 by construction and every other row reads as the fraction of
reference performance retained (Table~\ref{tab:robust}).

The lowest pooled environmental-condition mean was 76.6\% of the
pooled clear-baseline mean, occurring under aperture occlusion; mist
and dust retained 95.9\% and 92.6\%, respectively. Because target and
distance composition differ between conditions, these pooled ratios
are descriptive rather than causal estimates of environmental effect.
No scan across the full 2\,025 lost the forward sector entirely. By
target configuration, the separately normalized index ranges from
90.1 for the retroreflective panel to 63.9 for crinkled foil.

A diagnostic gate was positive when at least $k=3$ valid returns fell
within the forward sector at $\le$1.10\,m. It activated in all 1\,878
target-present scans, but also in all 147 no-target control scans
because in-gate walls and furniture were retained. These data therefore
confirm persistence of nearby forward-sector returns, not detection of
the labelled target, and neither precision nor target-level recall is
claimed. The operational stop proposal uses a separate 0.50\,m close
zone.

\subsection{Attitude, Vibration, and Observed Channel Continuity}
\label{sec:attitude}

\begin{table}[!t]
\caption{Attitude and LiDAR Yield, Representative Rows From Eight
Terrain Scenarios (mean $\pm$ SD)}
\label{tab:attitude}
\centering
\renewcommand{\arraystretch}{0.98}
\setlength{\tabcolsep}{3pt}
\resizebox{\columnwidth}{!}{%
\begin{tabular}{lccc}
\toprule
Scenario & Roll ($^{\circ}$) & Pitch ($^{\circ}$) & LiDAR valid (\%) \\
\midrule
Flat, stationary   & $0.00 \pm 3.34$  & $0.00 \pm 3.18$   & 68.2 \\
Static incline, $41^{\circ}$ nose-up & $-0.59 \pm 2.78$ &
$41.12 \pm 2.22$  & 73.6 \\
Static tilt, $47^{\circ}$ right & $-47.05 \pm 2.37$&
$1.58 \pm 1.69$   & 64.8 \\
Moving, flat       & $-6.77 \pm 5.05$ & $7.13 \pm 11.12$  & 71.6 \\
Moving, inclined   & $-2.82 \pm 8.24$ & $-45.31 \pm 5.67$ & 70.7 \\
Moving, rough      & $-8.96 \pm 62.42$& $7.99 \pm 29.85$  & \textbf{74.8} \\
\bottomrule
\end{tabular}%
}
\end{table}

Wheelchairs operate on ramps, thresholds, and uneven ground, so the
inertial channel was characterized across a flat baseline, four static
tilts, and three traversals; an additional disturbed traversal captured
the dropout described below. Roll and pitch were calculated from the
calibrated accelerometer vector, with a post-hoc flat-reference offset
correction, rather than from a complementary filter
(Table~\ref{tab:attitude} gives representative rows). On the four
static tilts the standard deviation of the principal tilted axis was
at most $2.4^{\circ}$ at indicated angles up to $47^{\circ}$. These
values quantify repeatability, not absolute angle accuracy, because no
independent angular reference was recorded. Under motion the estimates
broaden, and on rough terrain per-sample readings become unusable
(roll SD $62.4^{\circ}$); a temporal variance threshold would need
separate validation before use as a hazard detector.

The last column reports whole-scan LiDAR valid yield during each run,
a coarser metric than the forward-sector index of
Section~\ref{sec:degradation} because these runs have no fixed target
at a known bearing. Yield stayed within 64.8 to 74.8\% across the
reported scenarios, including rough-terrain traversal, and did not
decrease during motion relative to the flat baseline. This
valid-return statistic remained stable while roll and pitch dispersed;
it does not by itself establish range accuracy or target coverage while
the scan plane tilts.

This separation was observed during one
rough-terrain run, vibration intermittently broke contact on the
inertial unit's I$^2$C jumper wiring: only \textbf{41 of 100} expected
samples were recorded, with a maximum inter-sample gap of
\textbf{2.06\,s} against $\le$0.62\,s in every other run. Over the
same interval the logger recorded 115 LiDAR scans and the expected
camera samples. The gap exceeds the proposed 500\,ms channel timeout,
although the monitor and alert response were not implemented during
this run. The defect was traced to jumper wiring and remediated with
soldered connections and external pull-ups. The episode provides one
observed non-simultaneous channel failure, rather than proof of general
independence or a tested fail-safe response.

\section{Integrated System Checks and Discussion}
\label{sec:results}

\begin{figure}[!t]
\centering
\includegraphics[width=0.95\columnwidth]{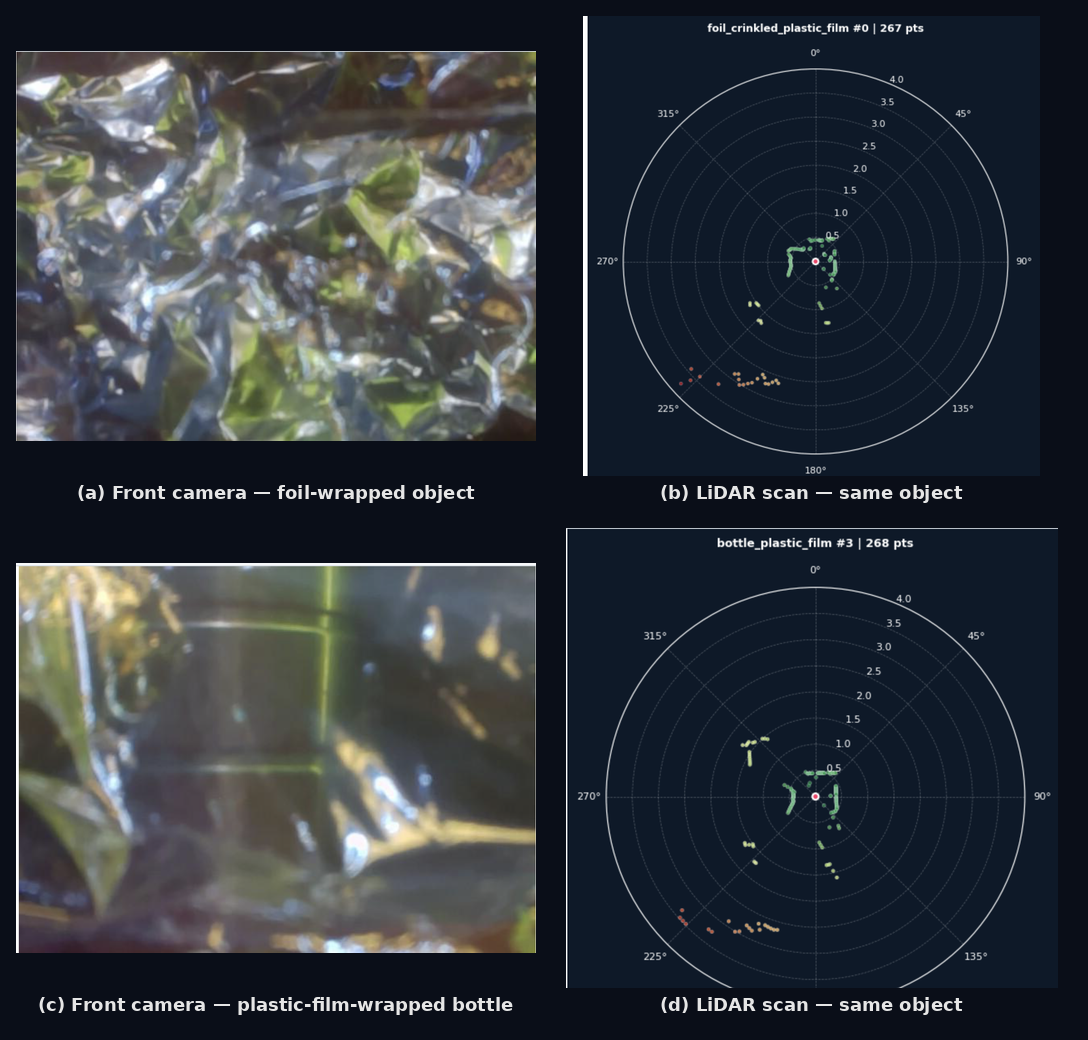}
\caption{Camera and LiDAR responses to optically adverse objects.
(a) Front camera frame of a foil-wrapped object showing severe
specular glare; (b) a corresponding 267-point LiDAR scan; (c) a camera
frame with low-contrast blur; and (d) a corresponding 268-point LiDAR
scan. The images illustrate complementary observations; target-level
detection and exact inter-channel synchronization are not claimed.}
\label{fig:camlidar}
\end{figure}

\subsection{Integrated Operation}

Section~\ref{sec:sensing} characterizes individual channels. The
following observations were collected with actuation, streaming, and
sensing running concurrently. They are integration checks rather than
an instrumented field study with a defined route and statistical
power.

End-to-end command latency, from emission at the dashboard to
actuation at the safety controller, was observed to fall in a 200 to
350\,ms band over a 2.4\,GHz link with concurrent camera streams
active. This is an \emph{observed operating band}, not an instrumented
distribution: it was not captured with a hardware-anchored
timestamping protocol, and no percentile statistics are claimed. The band
is high enough to impose the operator load characterized in
\cite{latency2024}, but remains within the sub-second envelope
reported for comparable caregiver-mediated frameworks
\cite{aiotframework2025}. The targeted use case is low-speed
($<$2\,km/h) operation. The front camera held a stable
15\,fps at 640$\times$480 with observable latency near 400 to
600\,ms; concurrent full-resolution streaming of all four cameras
saturated the wireless bandwidth, which is what motivates the
prioritized streaming policy of Section~\ref{sec:platform}.

\subsection{Complementary Failure of Camera and LiDAR}

Reflective and translucent material produced specular glare and
low-contrast blur in the camera views
(Fig.~\ref{fig:camlidar}(a),(c)), while corresponding LiDAR scans
retained nearby clusters (Fig.~\ref{fig:camlidar}(b),(d)). This
qualitative observation is consistent with complementary LiDAR and
vision failure modes reported previously \cite{multisensor2025,
lidar_denoise2020}; it is not a quantified camera-detection test.

\subsection{Privacy-Preserving Route Protocol}

\begin{figure}[!t]
\centering
\includegraphics[width=\columnwidth]{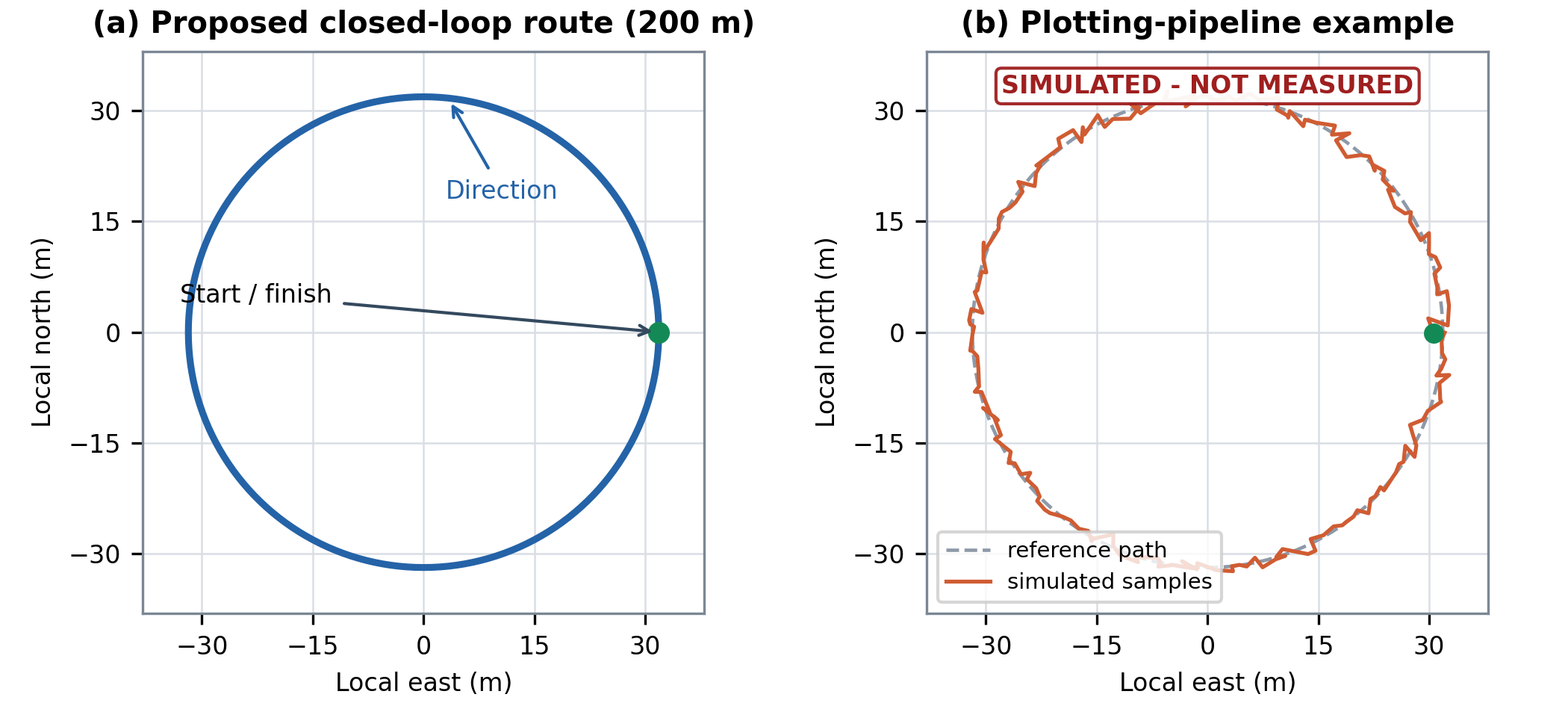}
\caption{Coordinate-free template for a proposed 200\,m closed-loop
field trial. (a) The route is expressed in local east/north coordinates
to avoid publishing a location. (b) Simulated position samples verify
the plotting pipeline and are explicitly not measured vehicle data.
This panel must be replaced with logged GNSS samples before it can
serve as field-validation evidence.}
\label{fig:routeplan}
\end{figure}

Fig.~\ref{fig:routeplan} defines a practical privacy-preserving field
protocol: record GNSS position, speed, command latency, battery voltage,
attitude, packet loss, and safety events on a common clock for three
200\,m laps. Convert latitude and longitude to local east/north offsets
before publication and report route completion, mean $\pm$ SD speed and
latency, and intervention counts. No GNSS log was found for the present
study, so the plotted samples are a transparent template rather than a
field result.

\subsection{Cost and Positioning}

The complete bill of materials totals approximately US\$335 at
indicative online retail pricing (April 2026), excluding the host
wheelchair. The LiDAR (US\$90) and the primary controller with storage
(US\$90) account for 53.7\% of the total; four camera channels
(US\$64), battery (US\$25), motor drive and motors (US\$21), safety
controller, inertial unit and servo (US\$17), and chassis, wiring, and
emergency stop (US\$28) make up the rest. As
Table~\ref{tab:related} shows, AVD does not match the millimetre
docking precision of \cite{lidarwheelchair2024}, and its detection
result is reported on deliberately adverse, uncleared scenes rather
than a prepared range. AVD combines live multi-camera teleoperation
with an explicit fault-response design and sensor-degradation
characterization.

\subsection{Limitations}
\label{sec:limitations}

\textbf{Joystick-bypass validation is the single most important
unproven claim in this work.} All results here were collected on a
proof-of-concept chassis driven by hobby-grade motors through a
discrete H-bridge. The reverse-engineered interface was therefore
validated only as a signal source into a discrete driver, \emph{not}
into the proprietary power module of a commercial powered wheelchair.
The central feasibility assumptions, that a commercial drive
controller accepts externally synthesized joystick-equivalent signals,
that its interlocks neither suppress nor fault on them, and that its
fault-detection logic tolerates the substitution, remain
\textbf{untested on real hardware}. Until a bench validation on one
commercial analog-interface chair is performed, this should be read as
a proposed and partially prototyped methodology rather than a
demonstrated capability. The power-stage equivalence argument of
Section~\ref{sec:platform} does not extend to this claim.

Four further limitations bound the study. Target-level detection
precision and recall are not established because the control scenes
were not background-cleared. The
robustness campaign used one sensor unit at one indoor site, so the
absolute yields and relative ratios remain specific to the sampled
scene compositions. End-to-end
latency is an observed band rather than an instrumented distribution,
no formal trials with persons with disabilities have been conducted
\cite{paruserstudy2024}, and the proposed 200\,m repeated field route
has not yet been run. Finally, the planar LiDAR senses one horizontal slice
and is structurally blind to drop-offs below and overhangs above it,
so the obstacle map must not be read as a clearance guarantee.

\emph{Future work.} The immediate next milestones are commercial-chair
bench validation and three instrumented laps of the 200\,m route in
Fig.~\ref{fig:routeplan}, followed by implementation and fault-injection
testing of the specified safety-controller policies.

\section{Conclusion}

AVD is presented as a safety-oriented teleoperation architecture for
assistive mobility. Its core design choices are a static authority
hierarchy with stop-only sensing and a separate safety-controller
domain. The paper distinguishes measured channel behavior from
specified fault responses: the LiDAR degradation and acquisition data
are measured, one inertial dropout is logged, and the watchdog,
degraded-mode, and supply-isolation policies await implementation or
fault-injection testing. The route protocol in Fig.~\ref{fig:routeplan}
defines the next field-validation step without presenting simulated
positions as experimental results.

\section*{Code Availability}

Public prototype code is available for the
\href{https://github.com/NamaiBest/LiDAR-Based-Suggestive-Navigation}{YDLiDAR interface}
and the
\href{https://github.com/NamaiBest/MPU-6050-9250_Raspberry-Pi-5}{inertial interface}.
A complete archival AVD control-stack release is pending.

\begingroup
\renewcommand{\footnotesize}{\scriptsize}
\bibliographystyle{IEEEtran}
\bibliography{references}

@misc{who2024,
  author       = {{World Health Organization}},
  title        = {Assistive Technology: Fact Sheet},
  year         = {2024},
  month        = jan,
  url          = {https://www.who.int/news-room/fact-sheets/detail/assistive-technology},
  note         = {Accessed: Sep. 9, 2026}
}

@article{simpson2005,
  author  = {R. C. Simpson},
  title   = {Smart wheelchairs: {A} literature review},
  journal = {Journal of Rehabilitation Research and Development},
  volume  = {42},
  number  = {4},
  pages   = {423--436},
  year    = {2005},
  doi     = {10.1682/JRRD.2004.08.0101}
}

@article{leaman2017,
  author  = {J. Leaman and H. M. La},
  title   = {A comprehensive review of smart wheelchairs: Past, present,
             and future},
  journal = {IEEE Transactions on Human-Machine Systems},
  volume  = {47},
  number  = {4},
  pages   = {486--499},
  month   = aug,
  year    = {2017},
  doi     = {10.1109/THMS.2017.2706727}
}

@article{bourhis1996,
  author  = {G. Bourhis and P. Pino},
  title   = {Mobile robotics and mobility assistance for people with motor
             impairments: Rational justification for the {VAHM} project},
  journal = {IEEE Transactions on Rehabilitation Engineering},
  volume  = {4},
  number  = {1},
  pages   = {7--12},
  month   = mar,
  year    = {1996},
  doi     = {10.1109/86.486052}
}

@article{levine1999,
  author  = {S. Levine and D. Bell and L. Jaros and R. Simpson and
             Y. Koren and J. Borenstein},
  title   = {The {NavChair} assistive wheelchair navigation system},
  journal = {IEEE Transactions on Rehabilitation Engineering},
  volume  = {7},
  number  = {4},
  pages   = {443--451},
  month   = dec,
  year    = {1999},
  doi     = {10.1109/86.808948}
}

@article{garcia2023,
  author  = {J. C. Garc{\'i}a and others},
  title   = {Filling the gap between research and market: Portable
             architecture for an intelligent autonomous wheelchair},
  journal = {International Journal of Environmental Research and
             Public Health},
  volume  = {20},
  number  = {2},
  pages   = {1243},
  month   = jan,
  year    = {2023},
  doi     = {10.3390/ijerph20021243}
}

@article{headcontrol2025,
  author  = {A. Haddoun and D. Djabri and M. Saidani and M. Benbouzid},
  title   = {Development and evaluation of a head-controlled wheelchair
             system for users with severe motor impairments},
  journal = {MethodsX},
  volume  = {15},
  pages   = {103485},
  year    = {2025},
  doi     = {10.1016/j.mex.2025.103485}
}

@inproceedings{eyecontrol2015,
  author    = {S. N. Patel and V. Prakash},
  title     = {Autonomous camera based eye controlled wheelchair system
               using {Raspberry Pi}},
  booktitle = {Proc. IEEE Int. Conf. on Innovations in Information,
               Embedded and Communication Systems (ICIIECS)},
  pages     = {1--6},
  year      = {2015},
  doi       = {10.1109/ICIIECS.2015.7192876}
}

@article{handgesture2025,
  author  = {M. R. Huda and others},
  title   = {Developing a real-time hand-gesture recognition technique
             for wheelchair control},
  journal = {PLOS ONE},
  volume  = {20},
  number  = {4},
  pages   = {e0319996},
  month   = apr,
  year    = {2025},
  doi     = {10.1371/journal.pone.0319996}
}

@article{eyevoice2020,
  author  = {S. Anwer and A. Waris and H. Sultan and S. I. Butt and
             M. H. Zafar and M. Sarwar and I. K. Niazi and
             M. Shafique and A. N. Pujari},
  title   = {Eye and voice-controlled human machine interface system for
             wheelchairs using image gradient approach},
  journal = {Sensors},
  volume  = {20},
  number  = {19},
  pages   = {5510},
  year    = {2020},
  doi     = {10.3390/s20195510}
}

@article{voice2020,
  author  = {M. M. Abdulghani and K. M. Al-Aubidy and M. M. Ali and
             Q. J. Hamarsheh},
  title   = {Wheelchair neuro fuzzy control and tracking system based on
             voice recognition},
  journal = {Sensors},
  volume  = {20},
  number  = {10},
  pages   = {2872},
  year    = {2020},
  doi     = {10.3390/s20102872}
}

@article{bci2025,
  author  = {S. A. H. Hussain and I. Raza and S. A. Hussain and
             M. H. Jamal and T. Gulrez and A. Zia},
  title   = {A mental state aware brain computer interface for adaptive
             control of electric powered wheelchair},
  journal = {Scientific Reports},
  volume  = {15},
  number  = {1},
  year    = {2025},
  doi     = {10.1038/s41598-024-82252-7}
}

@article{emg2024,
  author  = {H. Iqbal and J. Zheng and R. Chai and S. Chandrasekaran},
  title   = {Electric powered wheelchair control using user-independent
             classification methods based on surface electromyography
             signals},
  journal = {Medical \& Biological Engineering \& Computing},
  volume  = {62},
  number  = {1},
  pages   = {167--182},
  year    = {2024},
  doi     = {10.1007/s11517-023-02921-z}
}

@article{multisensor2025,
  author  = {Y. Zhang and Y. Wang and C. Su and Y. Miao and T. Wei and
             Y. Feng and F. Chen and Z. Ying and S. Wang and X. Wang},
  title   = {Multi-sensor fusion-based intelligent auxiliary system of
             power wheelchairs for individuals with limbs disabilities:
             Design and implementation},
  journal = {Measurement},
  volume  = {257},
  pages   = {118573},
  year    = {2026},
  doi     = {10.1016/j.measurement.2025.118573}
}

@article{hsu2012,
  author  = {P.-E. Hsu and Y.-L. Hsu and K.-W. Chang and C. Geiser},
  title   = {Mobility Assistance Design of the Intelligent Robotic
             Wheelchair},
  journal = {International Journal of Advanced Robotic Systems},
  volume  = {9},
  number  = {6},
  year    = {2012},
  doi     = {10.5772/54819}
}

@article{sensorfusion2023,
  author  = {P. Theodorou and K. Tsiligkos and A. Meliones},
  title   = {Multi-sensor data fusion solutions for blind and visually
             impaired: Research and commercial navigation applications
             for indoor and outdoor spaces},
  journal = {Sensors},
  volume  = {23},
  number  = {12},
  pages   = {5411},
  month   = jun,
  year    = {2023},
  doi     = {10.3390/s23125411}
}

@article{deepfusion2024,
  author  = {A. Thakur and S. K. Mishra},
  title   = {An in-depth evaluation of deep learning-enabled adaptive
             approaches for detecting obstacles using sensor-fused data
             in autonomous vehicles},
  journal = {Engineering Applications of Artificial Intelligence},
  volume  = {133},
  pages   = {108550},
  year    = {2024},
  doi     = {10.1016/j.engappai.2024.108550}
}

@article{lidarslam2023,
  author  = {L. Mochurad and Y. Hladun and R. Tkachenko},
  title   = {An obstacle-finding approach for autonomous mobile robots
             using {2D LiDAR} data},
  journal = {Big Data and Cognitive Computing},
  volume  = {7},
  number  = {1},
  pages   = {43},
  year    = {2023},
  doi     = {10.3390/bdcc7010043}
}

@article{lidarwheelchair2024,
  author  = {X. Lei and C. Tang and X. Tang},
  title   = {High-precision docking of wheelchair/beds through {LIDAR}
             and visual information},
  journal = {Frontiers in Bioengineering and Biotechnology},
  volume  = {12},
  pages   = {1446512},
  month   = sep,
  year    = {2024},
  doi     = {10.3389/fbioe.2024.1446512}
}

@article{iotchair2022,
  author  = {J. Cui and L. Cui and Z. Huang and X. Li and F. Han},
  title   = {{IoT} wheelchair control system based on multi-mode sensing
             and human-machine interaction},
  journal = {Micromachines},
  volume  = {13},
  number  = {7},
  pages   = {1108},
  year    = {2022},
  doi     = {10.3390/mi13071108}
}

@misc{multimodalnav2026,
  author       = {M. A. Hossain and M. E. Hoque},
  title        = {Advancing assistive robotics: Multi-modal navigation and
                  biophysical monitoring for next-generation wheelchairs},
  howpublished = {arXiv preprint arXiv:2601.02766},
  year         = {2026},
  url          = {https://arxiv.org/abs/2601.02766}
}

@misc{conav2025,
  author       = {Y. Xu and Q. Wang and J. Lillie and V. Kamat and
                  C. Menassa and C. D'Souza},
  title        = {{CoNav Chair}: Development and evaluation of a shared
                  control-based wheelchair for the built environment},
  howpublished = {arXiv preprint arXiv:2507.11716},
  year         = {2025},
  url          = {https://arxiv.org/abs/2507.11716}
}

@article{latency2024,
  author  = {Y. Ye and others},
  title   = {Brain functional connectivity under teleoperation latency:
             a {fNIRS} study},
  journal = {Frontiers in Neuroscience},
  volume  = {18},
  month   = nov,
  year    = {2024},
  doi     = {10.3389/fnins.2024.1416719}
}

@inproceedings{tcpstream2015,
  author    = {S. Ando and Y. Hayashi and T. Mizuki and H. Sone},
  title     = {Basic study on the method for real-time video streaming
               with low latency and high bandwidth efficiency},
  booktitle = {Proc. IEEE 39th Annual Computer Software and Applications
               Conference (COMPSAC)},
  pages     = {79--82},
  year      = {2015},
  doi       = {10.1109/COMPSAC.2015.217}
}

@misc{aiotframework2025,
  author       = {S.-C. Sun and Y.-C. Tsai},
  title        = {A modular {AIoT} framework for low-latency real-time
                  robotic teleoperation in smart cities},
  howpublished = {arXiv preprint arXiv:2510.11421},
  month        = oct,
  year         = {2025},
  url          = {https://arxiv.org/abs/2510.11421}
}

@misc{jones1977,
  author       = {M. L. Jones and P. M. Jobson},
  title        = {Wheelchair Control Circuit},
  howpublished = {U.S. Patent 4,059,786, assigned to Possum Controls Ltd.},
  month        = nov,
  year         = {1977}
}

@misc{rnet_canbus,
  author       = {{PG Drives Technology (Curtiss-Wright)}},
  title        = {{R-net Control System: Multi-Module Powerchair Control}},
  howpublished = {Technical brochure; ReBus protocol over CAN bus},
  year         = {2020}
}

@article{lidar_denoise2020,
  author  = {R. Heinzler and F. Piewak and P. Schindler and W. Stork},
  title   = {{CNN}-based Lidar Point Cloud De-Noising in Adverse Weather},
  journal = {IEEE Robotics and Automation Letters},
  volume  = {5},
  number  = {2},
  pages   = {2514--2521},
  year    = {2020},
  doi     = {10.1109/LRA.2020.2972865}
}

@misc{teleopvideo2021,
  author       = {A. Schimpe and S. Hoffmann and F. Diermeyer},
  title        = {Adaptive Video Configuration and Bitrate Allocation for
                  Teleoperated Vehicles},
  howpublished = {arXiv preprint arXiv:2102.10898},
  year         = {2021}
}

@misc{smartwheelchair2018,
  author       = {B. Narin and M. Brian and W. D. Smart},
  title        = {A Critical Look at Smart Wheelchairs},
  howpublished = {arXiv preprint arXiv:1809.00291},
  year         = {2018}
}

@misc{paruserstudy2024,
  author       = {A. Nanavati and M. Pascher and V. Ranganeni and
                  E. K. Gordon and T. Kessler Faulkner and S. S. Srinivasa and
                  M. Cakmak and P. Alves-Oliveira and J. Gerken},
  title        = {Multiple Ways of Working with Users to Develop
                  Physically Assistive Robots},
  howpublished = {arXiv preprint arXiv:2403.00489},
  year         = {2024}
}

@article{avizienis2004,
  author  = {A. Avizienis and J.-C. Laprie and B. Randell and
             C. Landwehr},
  title   = {Basic Concepts and Taxonomy of Dependable and Secure
             Computing},
  journal = {IEEE Transactions on Dependable and Secure Computing},
  volume  = {1},
  number  = {1},
  pages   = {11--33},
  month   = jan,
  year    = {2004},
  doi     = {10.1109/TDSC.2004.2}
}

@misc{iec61508,
  author       = {{International Electrotechnical Commission}},
  title        = {{IEC 61508-1}: Functional Safety of
                  Electrical/Electronic/Programmable Electronic
                  Safety-Related Systems, Part 1: General Requirements},
  howpublished = {International Standard, 2nd ed.},
  address      = {Geneva, Switzerland},
  year         = {2010}
}

@misc{iso7176,
  author       = {{International Organization for Standardization}},
  title        = {{ISO 7176-14}: Wheelchairs, Part 14: Power and Control
                  Systems for Electrically Powered Wheelchairs and
                  Scooters, Requirements and Test Methods},
  howpublished = {International Standard},
  address      = {Geneva, Switzerland},
  year         = {2022}
}
\endgroup

\end{document}